\documentclass[11pt]{article}

\usepackage{newtxtext,newtxmath}
\usepackage[T1]{fontenc}
\usepackage{amsmath}
\usepackage{amsfonts}
\usepackage{bm}
\usepackage{graphicx}
\usepackage[top=1in, bottom=1in, right=1in, left=1in]{geometry}
\usepackage{setspace}
\usepackage{tikz}
\usetikzlibrary{arrows.meta,bending,positioning,fit,fadings} 
\usepackage[authordate, autocite=inline, backend=bibtex, natbib]{biblatex-chicago}
\usepackage{sectsty}
\allsectionsfont{\sffamily}
\usepackage{subcaption}
\DeclareCaptionFormat{custom}
{%
    \textsf{\textbf{#1#2}#3}
}
\usepackage{abstract}

\usepackage[hidelinks]{hyperref}
\usepackage{authblk}
\usepackage{tabularray}
\usepackage{lettrine}
\usepackage{placeins}
\usepackage{afterpage}
\usepackage{enumitem}
\usepackage{standalone}

\title{\textsf{\textbf{Measuring the Quality of Answers in Political Q\&As with Large Language Models}}}
\author{R. Michael Alvarez}
\author{Jacob Morrier\thanks{Corresponding author. Email address: \href{mailto:jmorrier@caltech.edu}{jmorrier@caltech.edu}.}}
\affil{Division of the Humanities and Social Sciences \\ California Institute of Technology}
\date{February 2025}

\begin{document}

\maketitle

\begin{abstract}
    This article proposes a new approach for assessing the quality of answers in political question-and-answer sessions. We measure the quality of an answer based on how easily and accurately it can be recognized in a random set of candidate answers given the question's text. This measure reflects the answer's relevance and depth of engagement with the question. Like semantic search, we can implement this approach by training a language model on the corpus of observed questions and answers without additional human-labeled data. We showcase and validate our methodology within the context of the Question Period in the Canadian House of Commons. Our analysis reveals that while some answers have a weak semantic connection to questions, hinting at some evasion or obfuscation, they are generally at least moderately relevant, far exceeding what we would expect from random replies. We also find a meaningful correlation between answer quality and the party affiliation of the members of Parliament asking the questions.
\end{abstract}

\bigskip


\clearpage

\section{Introduction}

\citet[p.~115]{bull_1994} asserts that ``in any democratic system, questions play an important role in political communication.'' The prevalence of question-and-answer (Q\&A) sessions in political settings supports this claim. Prominent examples include election debates, press conferences, and town hall meetings. Additionally, institutional practices, such as congressional hearings and parliamentary questions, provide formal venues for these direct exchanges. They are generally considered vital for the effective oversight of the executive branch.

The primary objective of Q\&As is to elicit informative answers from those questioned, though participants may have additional motives. These additional objectives can undermine the quality of replies, such as when political messaging takes precedence over substantive discussions. In this context, assessing answer quality is imperative, as it reflects the efficacy of the political institutions espousing the Q\&A format.

However, measuring the quality of answers in political Q\&As is challenging. A key difficulty lies in defining and operationalizing answer quality, a complex and multifaceted concept. For example, \citet{bull_mayer_1993} identified 30 tactics to elude questions in political interviews. Furthermore, manually labeling large datasets with thousands of exchanges or more is prohibitively expensive, labor-intensive, and prone to human error, posing problems for human labeling of entire corpora and estimation with supervised learning.

Considering these challenges, we propose a novel approach for assessing the quality of answers in political Q\&As. We measure the quality of an answer based on how easily and accurately it can be recognized from a set of candidate answers based on the text of the question. This measure reflects the reply's relevance and depth of engagement with the question. We estimate it in a self-supervised way by training a language model on the corpus of observed questions and answers without additional human-labeled data. This is due to the strong parallel between our operationalization of answer quality and semantic search, a core task in natural language processing (NLP) that involves retrieving from a database the information most relevant to a query \citep{INR-032}. Rather than using these question-answering systems to identify the most relevant answer from a set of candidates, we propose repurposing them to gauge the quality of observed answers.

We showcase our methodology using the Question Period (QP) in the Canadian House of Commons as a case study. The House of Commons is the Parliament of Canada's lower chamber, where the Prime Minister and federal Cabinet ministers sit. QP takes place for 45 minutes every day the House is in session, drawing considerable attention from the media and public. During QP, members of Parliament (MPs), primarily those from opposition parties, can seek information from the government. It is one of the few moments in Parliament when the opposition sets the agenda.

We analyze the 58,343 exchanges that occurred during QP from 2006 to 2021. Our analysis reveals that while some answers are only loosely related to questions, suggesting some evasion or obfuscation, most are at least moderately relevant, surpassing what we expect from random replies. Upon measuring answer quality, we investigate its correlates, focusing on its relationship with power dynamics in the House of Commons. We observe a significant correlation between the quality of answers and the party affiliation of the MP asking questions, with questions from government backbenchers, MPs from third parties (as opposed to the official opposition), and those from parties ideologically closer to the government tending to receive more relevant answers.

Our work engages with a theoretical debate on the nature of QP, mirroring the tension between the deliberative and strategic approaches in comparative theories of parliamentary debates \citep{bachtinger_2014, whyte_2019}. While QP's official purpose is to allow MPs to seek the information necessary for effectively fulfilling their oversight responsibilities, some argue that it tends to devolve into a performative exercise, during which questions and answers are used for political messaging rather than for holding the government accountable. Our study assesses the extent to which and under what circumstances QP fulfills its intended function of enabling effective government oversight.

This paper is structured as follows. First, we situate our research within the existing literature. Next, we describe the origins and functions of QP. Building on this information, we operationalize answer quality and outline the data and methodology used to implement our proposed methodology. Then, we formulate and test hypotheses regarding the relationship between our measure of answer quality and relevant variables. Last, we discuss our methodology's broader applicability.

\section{Related Literature}

This paper is related to three bodies of academic literature, one substantive and two methodological.

Substantively, this study contributes to the literature on parliamentary questions, particularly QP. We build on recent research investigating the use of avoidance tactics by government ministers \citep{RASIAH2010664, Bull_Strawson_2019, kukec_2022} and also work on evasion tactics in political interviews \citep{bull_mayer_1993, bull_1994, bull_1998, bull_2000, bull_2004, Bull_Strawson_2019, Waddle_Bull_2020}. We extend this line of inquiry to the Canadian context, where the quality of answers in QP has received limited attention, except for \citet{whyte_2019}'s work. Unlike previous studies that relied primarily on human annotation and basic NLP, we employ advanced computational methods.

Methodologically, we contribute by integrating tools from NLP into political science by harnessing semantic search to address substantive questions. Our model builds on sentence embeddings from a variant of Bidirectional Encoder Representations from Transformers (BERT), known as ``Sentence-BERT'' \citep{reimers2019sentencebert}. Political methodologists have widely adopted BERT and its variants \citep[e.g.,][]{Bestvater_Monroe_2023,Wang_2023,Widmann_Wich_2023,Laurer_van_Atteveldt_Casas_Welbers_2024}. Unlike standard word embeddings, which provide context-agnostic representations, Sentence-BERT embeddings capture the meaning of tokens within their specific context. They have consistently performed well in semantic search. Although generative models are beginning to outperform BERT in general-purpose text embedding tasks, BERT remains a cornerstone of NLP due to its reliability and ease of deployment.

Finally, this paper engages with the literature on predictive accuracy as a substantive measure of interest in political science, an idea originating from research on political polarization \citep{Peterson_Spirling_2018, Goet_2019}. We build on the concept of using predictive accuracy as a substantive metric by proposing to measure the relevance of answers by how easily and accurately they can be identified based on the text of the question.

\section{Institutional Context}

QP takes place during 45 minutes each day the House of Commons is in session \citep[ch.~11]{HOC_PNP}. Each question and answer is limited to 35 seconds, resulting in a rapid and lively exchange. The session begins with the Speaker inviting the Leader of the Opposition to ask a series of questions, typically directed at the Prime Minister. Subsequent questions follow a predetermined order reflecting the parties' representation in the House. Although government backbenchers and independent MPs occasionally participate, most questions come from members of officially recognized opposition parties. Party whips coordinate their caucus's participation and submit a suggested speaking order to the Speaker's office. The government chooses which minister or parliamentary secretary will respond to each question.

QP serves two functions: it acts as an accountability and oversight mechanism and a platform for political communication. QP's official purpose is for MPs to seek information from the government, setting it apart from other parliamentary debates and any sequence of loosely related speeches. Its origins lie in the principle of responsible government \citep[ch.~2]{HOC_PNP}. Under this constitutional convention, the government must retain the confidence of a majority of its members to stay in power. Concretely, this means that some ``confidence motions'' must be adopted; otherwise, the government must resign or request the Parliament's dissolution. Regular opportunities for dialogue between the government and the opposition also contribute to upholding responsible government. QP is arguably the most prominent of these mechanisms and one of the few moments in Parliament when opposition parties set the agenda.

QP is also a platform for political communication, allowing participants to reach a wider audience. Given the intense media and public attention it garners, QP presents politicians with a valuable opportunity to enhance their public profile, cast their opponents in a negative light, and strengthen their public support. In practice, opposition MPs use QP to gather the information necessary for effective oversight and to ``score political points'' by challenging the government, providing commentary, and proposing alternatives. Similarly, the government uses this time not only to respond to the opposition's questions but also to counter criticisms, defend its policies, and highlight its achievements.

QP's role as a political communication platform is not inherently incompatible with its function as an accountability and oversight mechanism. For example, the fear of public criticism and embarrassment can drive government ministers to engage seriously with questions \citep{Kernaghan_1979}. It is particularly significant for a majority government since the opposition does not threaten its survival. QP's public nature can also undermine its effectiveness as an accountability and oversight mechanism. For example, it might nudge the government to avoid providing direct answers when they might reveal embarrassing truths. Also, as opposition parties and the government prioritize scoring political points, QP can devolve into a largely performative exercise. Finally, government backbenchers often pose questions not to gather information but to cast a favorable light on the government's achievements.

In sum, QP functions both as a mechanism for accountability and oversight and as a platform for political communication. These roles interact in complex ways, highlighting the difficulties in assessing the quality of answers in QP. Ministers are unlikely to provide unadorned answers, frequently using them to convey political messages. Also, opposition MPs may ask questions intended for purposes other than seeking information. We must be able to look past these political tactics to assess whether answers are meaningful and thereby contribute to holding the government accountable, as this is an explicit desideratum of responsible government.

\section{Operationalization of Answer Quality}

As described above, the formal purpose of QP is to enable MPs to seek information from the government. We wish to assess whether and to what extent answers fulfill this function. To this end, answers must be directly relevant to the questions and engage substantially with the issues raised \citep{maricut-akbik_2021}. Accordingly, we propose assessing the quality of answers based on the depth of their engagement with and their relevance to questions. In this framework, more relevant answers reflect a more effective QP.

Drawing a parallel with semantic search, we measure an answer's relevance based on how easily and accurately it can be recognized from a set of candidates randomly drawn from the corpus of observed answers based on the question's text. This approach builds on the premise that relevant answers are more easily linked to questions than irrelevant ones, as they are uniquely tailored to questions, making them stand out as the most probable choice. In contrast, less relevant answers are more difficult to connect, with the least pertinent so vague or generic that they could apply to many questions, making other answers appear just as, if not more, likely.

Our measure aligns with semantic similarity, a core concept in NLP based on the ``distributional hypothesis,'' which asserts that words and sentences occurring in similar contexts tend to have similar meanings \citep[p.~101]{jurafsky_martin_2024}. Accordingly, we can measure the relevance of a word or sentence to a given context by its likelihood of occurring in it. We typically learn these similarity relationships through ``fill in the blank'' tasks, in which models predict the word or sentence appearing in a given context.

This approach addresses many challenges inherent in operationalizing answer quality. For instance, we adopt a pragmatic perspective on relevance, focusing on the functional relationship between questions and answers rather than on a fixed and inherently partial list of characteristics. We can computationally implement our conception of answer quality without human-labeled data, fostering efficiency. Finally, following \citet{bull_1994}, we conceptualize answers as existing on a continuum.

Admittedly, our criterion considers a single dimension of answer quality. This requirement alone does not guarantee a satisfactory answer. For example, a reply that rephrases the question without adding new information may meet this standard. However, we can confidently classify any response that fails to meet this minimal requirement as low quality. In this case, speakers effectively talk past each other. Accordingly, our approach allows us to detect deliberate efforts to deflect attention and obscure the essence of the discussion.\footnote{Effective diversions are characterized by the absence of semantic ties to the question, ensuring that the answer does not unintentionally reinforce the topic it means to avoid. Like the phrase ``Don't think of an elephant!'' inevitably triggers the thought of an elephant, a reply with even a slight semantic tie to the question risks failing as a genuine diversion \citep{lakoff_2014}.} In sum, while this criterion is a necessary condition for high-quality answers, it is not sufficient and, thus, may overestimate answer quality. A more comprehensive assessment of answer quality may also consider factual accuracy, civility, clarity, conciseness, and comprehensiveness.

Besides, confounding variables, such as the nature of the questions, can affect our assessment of answer quality. All else equal, an easily distinguishable answer is more relevant than one that is difficult to link to the question. However, not all questions are equally easy to answer and engage with. For example, a lengthy preamble covering multiple topics precedes many poorly framed questions \citep{harperland}. These questions offer multiple rebuttal angles and make it harder for responders to engage comprehensively with the question. In this case, our criterion requires a high-quality answer to address as many of the topics raised in the preamble as possible, which is a reasonable expectation.

Nevertheless, the relationship between the nature of the questions and our measure of the quality of the answer is indirect. Thus, we cannot regard the latter as equally reflecting a notion of question quality. While some questions are inherently more challenging to answer, well-crafted questions do not guarantee high-quality answers. More fundamentally, the quality of questions should not be judged solely by the answers they elicit. No standard for assessing question quality parallels ours, based on desirable traits consistent with the QP's stated purpose.\footnote{A possible exception is that the question must pertain to the federal government's responsibilities.} Ultimately, this hypothetical relationship between the nature of questions and our measure of answer quality does not prevent meaningful comparisons of answer quality across variables, provided that the nature of questions is consistent across them.

In conclusion, we reiterate that prompting and supplying insightful answers is only one of the many goals pursued by participants during QP. Tensions can arise between these objectives, potentially undermining the relevance of answers. In this context, the incentives of both the government and the opposition parties shape the quality of answers. Nonetheless, it remains an essential feature to consider, as it reflects the caliber of the discussion and QP's efficacy as a mechanism for accountability and oversight.

\section{Methodology}

\subsection{Network Architecture}

We implement our approach for assessing answer quality in a self-supervised manner. We train an artificial neural network to identify which answer from a random set of candidates corresponds to each question based on their respective texts. This model is trained on the corpus of observed questions and answers without additional human-labeled data. It belongs to the general class of semantic search models, which analyze the meaning of a query and seek to return the most relevant value from a database.

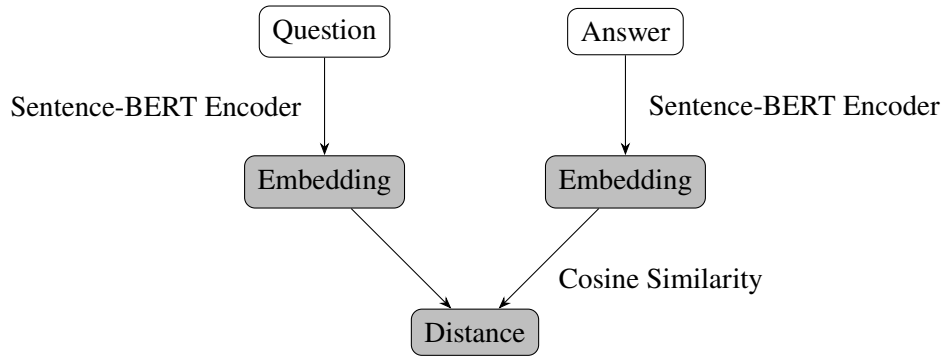
\begin{figure}[!p]
    \centering
        \begin{tikzpicture}
        \draw (0,2) node[draw, inner sep=5pt, rounded corners, fill=gray!50] (D) {Similarity}
        (-2,4) node[draw, inner sep=5pt, rounded corners, fill=gray!50] (E1) {Question Embedding}
        (2,4) node[draw, inner sep=5pt, rounded corners, fill=gray!50] (E2) {Answer Embedding}
        (-2,6) node[draw, inner sep=5pt, rounded corners] (Q) {Question}
        (2,6) node[draw, inner sep=5pt, rounded corners] (A) {Answer};
        \draw[-Stealth] (Q) -- (E1) node[midway, left=5pt]{Sentence-BERT Encoder};
        \draw[-Stealth] (A) -- (E2) node[midway, right=5pt]{Sentence-BERT Encoder};
        \draw[-Stealth] (E1) -- (D);
        \draw[-Stealth] (E2) -- (D) node[midway, anchor=north west] {Cosine Similarity};
    \end{tikzpicture}
    \bigskip
    \caption{Architecture of Biencoders}
    \label{fig:biencoders}
\end{figure}

The network has three primary layers: (i) an input layer that receives a question-answer pair, (ii) two identical Sentence-BERT encoders that independently process each input to produce dense numerical representations called embeddings, and (iii) the computation of a similarity metric between the embeddings. This architecture, known as a biencoder, is represented in Figure \ref{fig:biencoders}.

\begin{figure}[!p]
    \centering
    \makebox[\linewidth]{\begin{tikzpicture}
        \draw[opacity=0.25, rounded corners] (-3,1) rectangle (15,5);
        \node[anchor=south west, opacity=0.25] at (-3,5.1) {Encoder Layer $\left(\times N\right)$};
        \draw (0,0) node[draw, inner sep=5pt, rounded corners] (1) {Token 1}
        (6,0) node[draw, inner sep=5pt, rounded corners] (2) {Token 2}
        (9,0) node {...}
        (12,0) node[draw, inner sep=5pt, rounded corners] (N) {Token N}
        (0.2,2.2) node[draw, inner sep=5pt, rounded corners, fill=gray!50] {Self-Attention Head}
        (0.1,2.1) node[draw, inner sep=5pt, rounded corners, fill=gray!50] {Self-Attention Head}
        (0,2) node[draw, inner sep=5pt, rounded corners, fill=gray!50] (A) {Self-Attention Head}
        (0,4) node[draw, inner sep=5pt, rounded corners, fill=gray!50] (F) {Feed Forward};
        \begin{scope}[transparency group, opacity=0.25]
            \draw (6.2,2.2) node[draw, inner sep=5pt, rounded corners, fill=gray!50] {Self-Attention Head}
            (6.1,2.1) node[draw, inner sep=5pt, rounded corners, fill=gray!50] {Self-Attention Head}
            (6,2) node[draw, inner sep=5pt, rounded corners, fill=gray!50] (A2) {Self-Attention Head};
        \end{scope}
        \draw (6,4) node[draw, inner sep=5pt, rounded corners, fill=gray!50, opacity=0.25] (F2) {Feed Forward};
        \begin{scope}[transparency group, opacity=0.25]
            \draw(12.2,2.2) node[draw, inner sep=5pt, rounded corners, fill=gray!50] {Self-Attention Head}
            (12.1,2.1) node[draw, inner sep=5pt, rounded corners, fill=gray!50] {Self-Attention Head}
            (12,2) node[draw, inner sep=5pt, rounded corners, fill=gray!50] (A3) {Self-Attention Head};
        \end{scope}
        \draw (12,4) node[draw, inner sep=5pt, rounded corners, fill=gray!50, opacity=0.25] (F3) {Feed Forward};
        \draw (6,6) node[draw, inner sep=5pt, rounded corners, fill=gray!50] (P) {Pooling}
        (6,8) node[draw, inner sep=5pt, rounded corners, fill=gray!50] (SE) {Sentence Embedding};
        \draw[-Stealth] (1) -- (A);
        \draw[-Stealth] (1) to[out=180, in=180] (F);
        \draw[-Stealth] (A) -- (F);
        \draw[-Stealth] (F) -- (P);
        \draw[-Stealth, dashed] (2) -- (A);
        \draw[-Stealth, dashed] (N) -- (A);
        \draw[-Stealth, opacity=0.25] (2) -- (A2);
        \draw[-Stealth, opacity=0.25] (2) to[out=180, in=180] (F2);
        \draw[-Stealth, opacity=0.25] (A2) -- (F2);
        \draw[-Stealth, opacity=0.25] (F2) -- (P);
        \draw[-Stealth, dashed, opacity=0.25] (1) -- (A2);
        \draw[-Stealth, dashed, opacity=0.25] (N) -- (A2);
        \draw[-Stealth, opacity=0.25] (N) -- (A3);
        \draw[-Stealth, opacity=0.25] (N) to[out=0, in=0] (F3);
        \draw[-Stealth, opacity=0.25] (A3) -- (F3);
        \draw[-Stealth, opacity=0.25] (F3) -- (P);
        \draw[-Stealth, dashed, opacity=0.25] (1) -- (A3);
        \draw[-Stealth, dashed, opacity=0.25] (2) -- (A3);
        \draw[-Stealth] (P) -- (SE);
    \end{tikzpicture}}
    \bigskip
    \caption{Architecture of Sentence-BERT Encoders}
    \label{fig:BERT}
\end{figure}

Question and answer embeddings are generated by ``Sentence-BERT'' \citep{devlin-etal-2019-bert, reimers2019sentencebert}. The encoder architecture is illustrated in Figure \ref{fig:BERT}, with the branch for Token 1 highlighted and the others dimmed. An encoder takes sentences or short paragraphs, represented as ordered sequences of tokens corresponding to words or word segments, as input. It maps each token to its value's embedding and concatenates it to positional embeddings that encode its relative location in the sequence. These embeddings go through many successive layers consisting of a multi-head self-attention mechanism and a feed-forward component. A self-attention head considers the current token's embedding and those of the surrounding tokens and outputs a weighted sum of these embeddings based on learnable weights. Within each layer, multiple heads operate in parallel. This mechanism's output is combined with the original token embedding and passed through a fully connected layer. Ultimately, the encoder outputs a numerical vector for each input token, which we average to get sentence embeddings.

We use the cosine similarity to measure the resemblance between question and answer embeddings. After training the model, it becomes our measure of the quality of an answer relative to the question. For reference, the cosine similarity is a measure of the angle between two numerical vectors $\bm{x}$ and $\bm{y} \in \mathbb{R}^{n}$:
\[
    \cos\left(\bm{x}, \bm{y}\right) = \frac{\bm{x} \cdot \bm{y}}{\left \lVert \bm{x} \right \rVert \left \lVert \bm{y} \right \rVert}.
\]
By construction, the value of the cosine similarity is between $-$1 and 1, with two parallel vectors having a cosine similarity of 1, two orthogonal vectors a cosine similarity of 0, and two opposing vectors a cosine similarity of $-$1.

\subsection{Training Objective}

We adopt the Multiple Negatives Ranking Loss as our training objective \citep{henderson2017efficientnaturallanguageresponse}. Given a batch of size $K$, questions are represented by the embeddings $\bm{X} = \left(\bm{x}_{1}, ..., \bm{x}_{K}\right)$ and the corresponding answers by the embeddings $\bm{Y} = \left(\bm{y}_{1}, ..., \bm{y}_{K}\right)$. The objective is to minimize the mean negative log probability of observing the correct answer from all candidate answers in the batch. It is mathematically equivalent to maximizing the likelihood of observing the correct answer but improves numerical stability during optimization.

We model the probability of observing an answer as a logistic regression, where the input features are the cosine similarities between the question's embedding and those of the candidate answers. In particular, the probability of observing the correct answer equals:
$$
    P\left(\bm{y}_{i} \mid \bm{x}_{i}\right) = \frac{\exp\left(\alpha \cos\left(\bm{x}_{i}, \bm{y}_{i}; \bm{\theta}\right)\right)}{\sum_{j = 1}^{K} \exp\left(\alpha \cos\left(\bm{x}_{i}, \bm{y}_{j}; \bm{\theta}\right)\right)},
$$
where $\alpha$ is a fixed scaling parameter and $\bm{\theta}$ are the Sentence-BERT encoder's parameters.

Overall, our training objective is to find the parameters $\bm{\theta}$ that minimize the following function:
\begin{equation*}
    \begin{split}
        \mathcal{J}\left(\bm{X}, \bm{Y}; \bm{\theta}\right) & = -\frac{1}{K} \sum_{i = 1}^{K} \log\left(P\left(\bm{y}_{i} \mid \bm{x}_{i}\right)\right) \\
        & = -\frac{1}{K} \sum_{i = 1}^{K} \log\left(\frac{\exp\left(\alpha \cos\left(\bm{x}_{i}, \bm{y}_{i}; \bm{\theta}\right)\right)}{\sum_{j = 1}^{K} \exp\left(\alpha \cos\left(\bm{x}_{i}, \bm{y}_{j}; \bm{\theta}\right)\right)}\right) \\
        & = -\frac{1}{K} \sum_{i = 1}^{K} \left[\alpha \cos\left(\bm{x}_{i}, \bm{y}_{i}; \bm{\theta}\right) - \log\left(\sum_{j = 1}^{K} \exp\left(\alpha \cos\left(\bm{x}_{i}, \bm{y}_{j}; \bm{\theta}\right)\right)\right)\right].
    \end{split}
\end{equation*}

This equation contains two terms: the first term in the sum reflects the similarity between the question's embedding and the correct answer's embedding, while the second term reflects the similarity between the question's embedding and those of all candidate answers. Consistent with our geometric interpretation, the loss function leads the model to assign a higher similarity with the question to the correct answer and a lower similarity to incorrect answers.

Our training objective treats questions and answers asymmetrically, using questions as anchors. We could use answers as anchors instead, training the model to recognize the question that each answer addresses. This approach differs from traditional semantic search, where answers seldom need to be linked to questions, and generates estimates of how deeply each answer engages with the question enough that we can identify the latter from the former. Nonetheless, this approach produces cosine similarity estimates similar to ours, with a correlation coefficient of 0.959 (cf., Online Supplementary Material).

\subsection{Transfer Learning}

Consistent with best practices in NLP, we train our model with transfer learning \citep{ruder-etal-2019-transfer, Laurer_van_Atteveldt_Casas_Welbers_2024}. Rather than starting the training process from scratch, we begin with estimates from a model trained for general tasks on a large corpus different from ours that we subsequently ``fine-tune'' to our specific needs. Specifically, we use the \texttt{multi-qa-mpnet-base-cos-v1} model from the \texttt{sentence-transformers} library, trained for two tasks: (i) missing token prediction by filling in randomly masked words in short sentences and paragraphs from a large corpus of general text, and (ii) semantic search on a dataset of 215 million question-answer pairs from various online forums. Among the question-answering models available in the \texttt{sentence-transformers} library, this model achieves the highest performance in semantic search over six benchmark datasets. We fine-tune the model on five percent of the exchanges in our corpus, reserving one percent as a validation set to optimize training hyperparameters. We use the remaining 94\% for inference.

Fine-tuning consists of further updating the pre-trained model's parameters to improve its performance on our specific corpus and task.\footnote{The Online Supplementary Material contains details on the performance of the pre-trained and fine-tuned models over the inference set.} This is desirable because politicians are unlikely to answer questions in QP the same way users of online forums would. Nevertheless, the pre-trained model encodes general language patterns and a general understanding of semantic search and question answering. This foundational knowledge can considerably accelerate the learning of domain-adapted embeddings at minimum cost, with numerous studies showing that fully custom models typically offer no systematic improvements over fine-tuned pre-trained models \citep[e.g.,][]{arslan_2021}. Also, we reserve most of our corpus for inference, so we lack sufficient data to train a high-quality model from scratch. Finally, using a pre-trained model is an implicit form of regularization, reducing overfitting and improving generalizability.

Alternatively, we could use embeddings from the pre-trained model without any fine-tuning. This approach poses a risk: the model might ascribe a poor quality to some answers, not due to their inherent irrelevance but because it is unaccustomed to how politicians reply in political Q\&As. In contrast, our goal is to assess how effectively a model familiar with the context can accurately match questions to their answers.

Admittedly, fine-tuning also comes with risks. One concern is that it may allow the model to learn evasion tactics and connect questions with their answers despite the poor quality of those answers. It would result in conservative estimates that artificially inflate answer quality. Indeed, if we ignored evasion tactics, answers previously deemed relevant may no longer be, while those initially considered irrelevant may become even more so. However, pure evasion will tend to result in replies unrelated to the original question. To draw an analogy, prisoners do not escape by remaining in their cell or even relocating to another but by leaving the prison grounds entirely, with their success depending on the guards losing track of their whereabouts. It raises doubts about the model's ability to effectively learn such behavior, especially given its exposure to a limited number of question-answer pairs during fine-tuning, with only a portion involving evasion. It also stresses the paradoxical nature of ``predictable evasion'': an evasion strategy that we can easily associate with the question fails in its purpose.

To be sure, we compared the cosine similarity estimates between the pre-trained and fine-tuned models. We found a strong correlation between estimates from both models, with a coefficient of 0.744.\footnote{A scatterplot illustrating this correlation is enclosed in the Online Supplementary Material.} On average, question-answer pairs with low cosine similarity estimates in the pre-trained model have even lower estimates in the fine-tuned model, and pairs with the highest cosine similarity have similar cosine similarity estimates in both models. This implies that fine-tuning enhances the model's ability to distinguish between relevant and irrelevant pairs without driving a general rise in cosine similarity estimates. Furthermore, substantive results from the pre-trained model provided in the Online Supplementary Material are consistent with those from the fine-tuned model.

\subsection{Mean Reciprocal Rank}

We optimize the model's training hyperparameters based on its performance on the validation set as measured by the mean reciprocal rank (MRR), a standard metric for evaluating the accuracy of information retrieval systems \citep[p.~311]{jurafsky_martin_2024}. We compute the MRR by averaging across all questions in the validation set the reciprocal of the rank of the correct answer among all candidate answers in the validation set, ordered by decreasing similarity to the query:
\[
    \text{MRR} = \frac{1}{\left|Q\right|} \sum_{i = 1}^{\left|Q\right|} \frac{1}{\text{Rank}_{i}}.
\]
It is standard to set all ranks below a certain threshold to zero. Accordingly, we truncate all ranks below ten. The Online Supplementary Material contains MRR values for various training hyperparameter values and lists the training hyperparameters we retained.

The MRR accounts for the fact that our model ranks potential answers based on their likelihood of being correct. It aligns with our objective that correct answers attain the highest possible rank among candidate answers. In contrast, classification metrics like precision and recall aggregate all answers above a certain rank or threshold, such as all those above the tenth rank, treating cases in which the correct answer ranks first the same as those in which it ranks ninth. We can address this limitation by considering precision and recall for different thresholds.

\afterpage{\FloatBarrier}

\section{Data}

\begin{figure}[!p]
    \centering
    \includegraphics{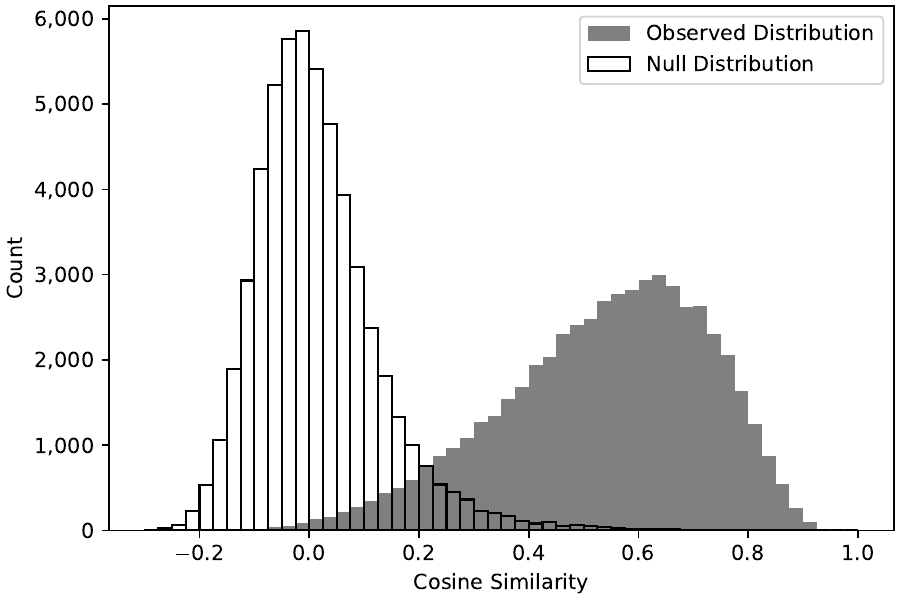}
    \caption{Distribution of the Cosine Similarity Between Questions and Answers}
    \label{fig:distribution}
\end{figure}

\begin{figure}[!p]
    \centering
    \begin{subfigure}[t]{0.475\textwidth}
        \centering
        \includegraphics[width=\linewidth]{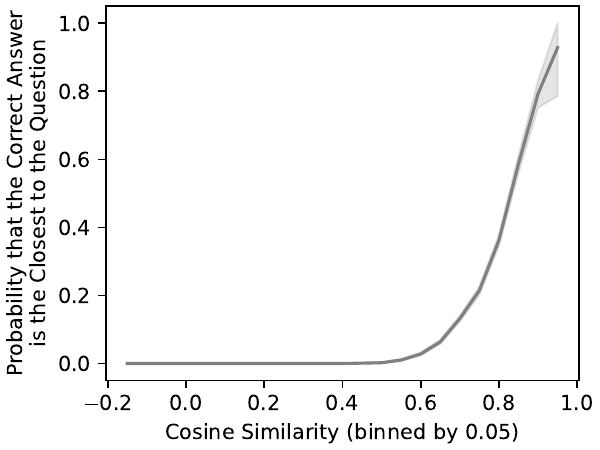}
        \caption{Probability That the Correct Answer is Closest to the Question by Cosine Similarity Between Questions and Answers}
    \end{subfigure}%
    \hfill
    \begin{subfigure}[t]{0.475\textwidth}
        \centering
        \includegraphics[width=\linewidth]{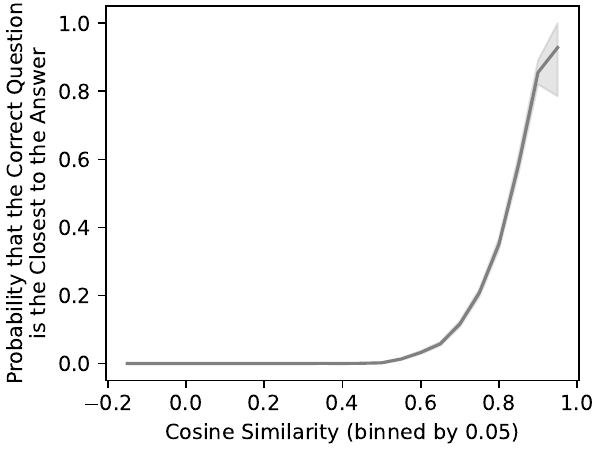}
        \caption{Probability That the Correct Question is Closest to the Answer by Cosine Similarity Between Questions and Answers}
    \end{subfigure}
    
    \bigskip
    
    \begin{subfigure}[t]{0.475\textwidth}
        \centering
        \includegraphics[width=\linewidth]{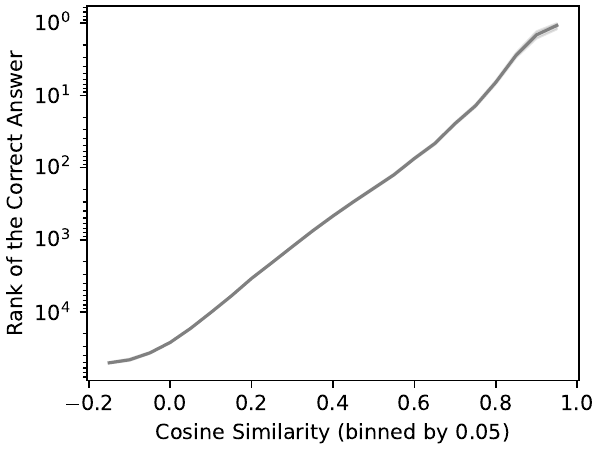}
        \caption{Rank of the Correct Answer by Cosine Similarity Between Questions and Answers}
    \end{subfigure}%
    \hfill
    \begin{subfigure}[t]{0.475\textwidth}
        \centering
        \includegraphics[width=\linewidth]{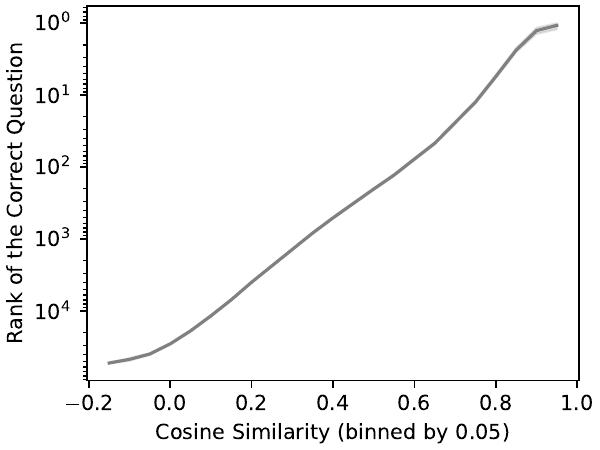}
        \caption{Rank of the Correct Question by Cosine Similarity Between Questions and Answers}
    \end{subfigure}
    \caption{Validity of the Cosine Similarity Between Questions and Answers}
    \label{fig:validity}
\end{figure}

\begin{table}[p]
    \centering
    \caption{Five Exchanges with the Lowest Cosine Similarity Between Questions and Answers}
    \label{tab:top_pairs}
    \scriptsize
    \begin{tblr}{colspec = {X|X|c}, columns = {valign = m}, rows = {rowsep = 5pt}, column{1-2} = {halign = l}, row{1} = {halign = c, font = \small}}
        \hline
        \hline
        \textbf{Question} & \textbf{Answer} & \textbf{Cosine Similarity} \\
        \hline
        Mr. Speaker, the Prime Minister confirmed yesterday that Justice Grenier had the full cooperation of the federal government during his investigation. That is entirely untrue. The federal government sent a lawyer who added numerous interventions specifically to prevent careful examination of federal spending. Why did the government go to so much trouble to protect the Liberals, Conservatives and the NDP from the investigation? What are they all trying to hide? & Mr. Speaker, let us all take a look at what the new Government of Canada has achieved since coming to power. We recognized that Québeckers form a nation within a united Canada, we resolved the issue of Quebec's presence at UNESCO, we resolved the fiscal imbalance issue. This concrete and positive action demonstrates very clearly that, together, Québec and Canada are progressing just fine, thank you. & $-$0.1245 \\
        \hline
        Mr. Speaker, the current Prime Minister participated in a demonstration in 2012, when he gave his word to Aveos workers. He said, and I quote, ``It is such a shame that we have to demonstrate to ask the law and order government to obey the law.'' More recently, he said, ``It is not true that our best resources are in the ground somewhere. Our best resources are human resources.'' Is that how a prime minister keeps his word? & Mr. Speaker, I reiterate that, of course, the Government of Canada is pleased by Air Canada's announcement of its intention to purchase the Bombardier CSeries aircraft. It is a major advancement in aviation. I am certain that this addition to the Air Canada fleet will be of major benefit, both to that company and to Canada's aerospace sector across the country. & $-$0.1310 \\
        \hline
        Mr. Speaker, the Prime Minister went to China to launch free trade negotiations, but the Chinese regime had something else in mind, even though the Prime Minister did everything he could to appease China and speed up takeovers of Canadian companies by waiving security reviews. The Prime Minister clearly has zero credibility when it comes to China. How are Canadians supposed to trust this Prime Minister to act in their best interest? & Mr. Speaker, I am looking forward to answering my colleague's questions, but first I would like to congratulate the four new members who were elected last night and who will be joining us here. I also want to highlight the 24 people who stepped up across the country to put their names on ballots in the by-elections. All of us in this place know what it takes to put your name on a ballot. I congratulate all of them, and all of the volunteers who underpin the strength of our democracy. I again look forward to congratulating the four new members when they arrive in this House. This was a good day for Canada, and a good day for our democracy. & $-$0.1323 \\
        \hline
        Mr. Speaker, if the finance minister were listening to Canadians, he would know that families are getting ripped off at the bank, ripped off at the gas pump, ripped off by cellphone companies and ripped off on their cable bills. But the rip-off does not end there. The finance minister is personally ripping off taxpayers. He paid a friend \$200,000 for a 20 page speech. Does he even know that \$200,000 is the average family's income for three years? This is unjustifiable. He has no moral authority to talk about budgetary matters or anything else. Why does he not just resign? & Mr. Speaker, the hon. member would know, if he bothered to review the material, that the work done was extensive. It was done by two people over an extensive period of several months. It related to policy and communications and not as the member just suggested. It is plain that the member has not bothered to review the documentation which is publicly disclosed.	& $-$0.1485 \\
        \hline
        Mr. Speaker, it seems that ``plus ça change, plus c'est pareil.'' Over the past few months, MPs have spent hundreds of hours hearing witnesses and debating on how to fight climate change in Canada. However, it seems the Conservative government does not care if Bill C-30 is ever brought to the floor of the House. Mr. Speaker, I am asking you today to get a search warrant to see if we can find Bill C-30 and bring it back to the House because the government is not going to do it. I ask you, Mr. Speaker, if you can find it, get it back to the House so we can debate it, get it passed and fight climate change now. & Mr. Speaker, I did note the recommendation of the hon. member, that people call me on this issue. I am gathering from some recent press reports that they should be able to reach me without calling at all; I can just hear through mediums. & $-$0.1625 \\
        \hline
        \hline
    \end{tblr}
\end{table}

\begin{table}[p]
    \centering
    \caption{Five Exchanges with the Highest Cosine Similarity Between Questions and Answers}
    \label{tab:bottom_pairs}
    \scriptsize
    \begin{tblr}{colspec = {X|X|c}, columns = {valign = m}, rows = {rowsep = 5pt}, column{1-2} = {halign = l}, row{1} = {halign = c, font = \small}}
        \hline
        \hline
        \textbf{Question} & \textbf{Answer} & \textbf{Cosine Similarity} \\
        \hline
        Mr. Speaker, due to the efforts of our government and based on our tremendous respect for their service to our country, Canada's injured veterans may receive an average monthly benefit of between \$4,000 to \$6,000. These are supports that our injured veterans need and deserve. Can the Parliamentary Secretary to the Minister of Veterans Affairs please update this House on the benefits that our government provides to injured veterans and their families? & Mr. Speaker, I thank my hon. colleague from Wild Rose for the question and his hard work on this file. Indeed, the average monthly financial benefit that an injured veteran may be eligible for is between \$4,000 to \$6,000 a month, and in some cases injured veterans are receiving a total income that exceeds \$10,000 a month. Our government is committed to ensuring that our injured veterans and their families have the support they need and deserve. Unfortunately, the members opposite have voted against virtually every single initiative that our government has brought forward to help Canada's veterans. & 0.9542 \\
        \hline
        Mr. Speaker, Canadians gave our government a strong mandate to end the wasteful and ineffective long gun registry. My constituents have told me repeatedly that they want to see an end to this measure, which needlessly and unfairly targets law-abiding hunters, farmers and sport shooters. We see the long gun registry as no less than an attack on our way of life. Could the Minister of Public Safety please update the House on what our government is doing to address this important issue? & Mr. Speaker, I thank the member for the work that he has done on this important file. On May 2, Canadians gave the government a strong mandate to end the wasteful and ineffective long gun registry once and for all, and that is exactly what we are doing. Canadians across the country have called for this measure. For example, Michelle Vardy of the Georgian Bay Women's Outdoors Workshops and the Ontario Federation of Anglers and Hunters stated: As a woman, the long gun registry does not make me feel any safer or more secure. It is wasteful, ineffective and reduces funding to do real things. The 2 billion dollars that have already been spent would have been better used on programs like healthcare--- & 0.9507 \\
        \hline
        Mr. Speaker, the people of China and Burma are suffering terribly in the aftermath of two tragic natural disasters. Canada responded immediately with an initial \$2 million to help the people of Burma when the cyclone hit. The unparalleled devastation in Burma has brought donor countries together to aid the victims of this tragedy. Could the Minister of International Cooperation update the House on our government's commitment to the victims in Burma and China?. & Mr. Speaker, Canada is deeply saddened by the tragic loss of life and devastation resulting from the disasters in Burma and China. We share the concerns of all Canadians for the victims and their families. Today I am announcing that our government will match the contributions of Canadians to humanitarian organizations working in Burma and China. Let me assure all Canadians our government will do our share of the international effort and ensure that our help does get to the victims and their families. & 0.9452 \\
        \hline
        Mr. Speaker, Toronto police chief Mark Saunders revealed in December that 82\% of handguns involved in crimes were smuggled from the U. S. The minister of public safety had previously stated half of crime guns come from domestic sources. The statistics from when the minister was the chief in Toronto and carried a gun show the same picture as today: A very small percentage of firearms are from legal sources, while many crime guns are prohibited and from the United States. Could the minister table the source of his information that has now been proven incorrect? & Mr. Speaker, unfortunately my friend has some of his facts wrong. When I was the chief of police in Toronto, we had a firearms verification unit that traced the source of all handguns. During my tenure as chief for 10 years there, 70\% of the crime guns that we seized, handguns, were smuggled from the United States. The other 30\% were stolen or illegally diverted. The 50\% number actually came from Chief Saunders in his first public statement, but he has since, as a result of some investigations they have done into smuggling, come out with another number. I acknowledge the facts there, but the reality is guns--- & 0.9382 \\
        \hline
        Mr. Speaker, Canada is an attractive place for African countries that are drawn by its bilingualism, its economic opportunities and its many top-notch institutions of higher learning. Last week the Prime Minister and many of his ministers were in Africa to develop new business opportunities. Could the Prime Minister please update the House on the actions our government is taking to expand trade between Ethiopia and Canada? & Mr. Speaker, I thank the member for Longueuil—Charles-LeMoyne for her question and her hard work. Expanding and diversifying trade between Canada and fast-growing African economies is a priority for our government. Trade between Canada and Ethiopia totaled \$170 million in 2018. We announced that we will be entering into negotiations towards a foreign investment promotion and protection agreement with Ethiopia, which will help further increase trade and investments for businesses in both countries. & 0.9374 \\
        \hline
        \hline
    \end{tblr}
\end{table}

We analyze all exchanges that took place during QP from the 39\textsuperscript{th} to the 43\textsuperscript{rd} legislatures, spanning the fifteen years between the January 23, 2006 election and the September 20, 2021 election. It amounts 58,343 exchanges, each consisting of a question and an answer. Our analysis focuses on questions from members of the four parties that have held official status at some point during our period of interest: the Bloc Québécois (BQ), the Conservative Party (CPC), the Liberal Party (LPC), and the New Democratic Party (NDP).\footnote{A party must have a minimum of twelve MPs to attain official status.} We construct our dataset from the official English transcripts published on the House of Commons website, which include professional translations of the interventions made in French.

Figure \ref{fig:distribution} depicts the distribution of cosine similarity for all question-answer pairs in the inference set and a null distribution of the cosine similarity between randomly matched questions and answers. This distribution is approximately Gaussian with a notable negative skew, reflecting a larger share of answers having a relatively low level of germaneness to initial questions compared to the share of answers with an equally high degree of pertinence. The answers with the lowest degree of relevance are orthogonal to the questions. Finally, the observed distribution of cosine similarity differs markedly from the null distribution, implying that the observed answers are significantly more relevant than random replies. While this is a low standard, it ensures that the answers have at least a superficial relevance to the questions.

Figure \ref{fig:validity} shows that the cosine similarity between questions and answers captures two factors: (i) the likelihood that the embeddings of a given question-answer pair are closest to each other, and (ii) the ranking of the correct question or answer among all candidates. Additional figures in the Online Supplementary Material further confirm the consistency of the cosine similarity across legislatures and the party affiliation of the MPs asking questions. This evidence dispels concerns that the correlation between answer quality and variables of interest revealed below is merely an artifact of variations in model accuracy across these features.

To illustrate what our measure of answer quality captures, Tables \ref{tab:top_pairs} and \ref{tab:bottom_pairs} present the five question-answer pairs with the lowest and highest cosine similarity, respectively. The exchanges with the lowest cosine similarity exemplify how a minister may fail to answer a question adequately. For example, the first two exchanges in Table \ref{tab:top_pairs} show ministers deflecting by underlining the government's achievements on tangentially related topics. The third exchange underscores that not all irrelevant answers arise from manifest ill intent. The fourth answer superficially touches on the question using generic talking points. In the fifth exchange, the minister dismisses the question with humor. Finally, only the second question suffers from poor framing, confirming that low-quality answers do not stem from poorly framed questions.

In contrast, the five exchanges in Table \ref{tab:bottom_pairs} feature detailed and precise answers. Many of the questions in these exchanges are emblematic of ``planted questions.'' We expected these questions and answers to exhibit high similarity. It remains unclear how these questions contribute to effective government accountability, underscoring the difficulty of operationalizing answer quality given QP's multiple functions. Nevertheless, the fourth exchange demonstrates that even a high-quality answer can challenge the question's premises, indicating that our measure of answer quality does not merely reflect collusion between the questioner and the respondent.

\afterpage{\FloatBarrier}

\section{Validity Experiment}

\begin{figure}[!p]
    \centering
    \includegraphics{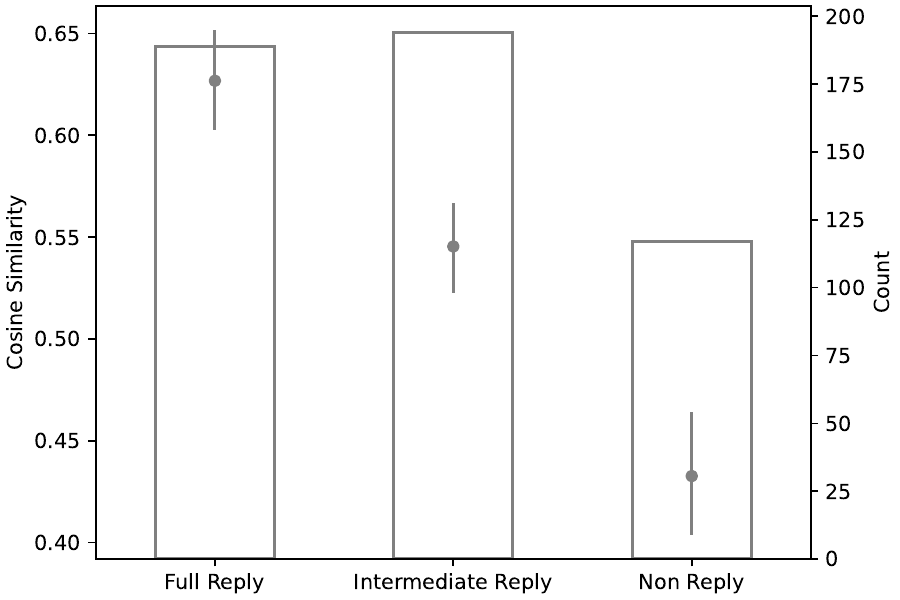}
    \caption{Average Cosine Similarity Between Questions and Answers and Count by Reply Category}
    \label{fig:validity_experiment}
\end{figure}

To assess the validity of our measure of answer quality, we compare it to labels produced by manually labeling a random sample of 500 question-answer pairs from the inference set. The labeling process follows the methodology used by \citet{Bates_et_al_2012} and \citet{Bull_Strawson_2019}. We classify each answer into one of three categories:
\begin{enumerate}[label=(\roman*)]
    \item \emph{Full reply:} An answer that thoroughly provides the requested information and/or conveys explicitly the government's position on the issue.

    \item \emph{Intermediate reply:} An answer that expresses the politician's views implicitly rather than explicitly and/or addresses only one part of a multi-pronged question.

    \item \emph{Non-reply:} An answer that fails to address the question, instead diverting to an unrelated topic, omitting the requested information, and/or withholding the government's stance on the issue.
\end{enumerate}

Our goal is not to show a perfect correlation between our measure and the benchmark labels, as they reflect different conceptions of answer quality. Furthermore, it would imply that our methodology is redundant. Nevertheless, we expect our measure to align moderately with existing conceptions of answer quality, especially since the traits they encompass generally make it easier to recognize answers. Accordingly, we expect full replies to receive, on average, higher quality estimates than intermediate replies, which, in turn, will receive higher average estimates than non-replies. If this holds, it will confirm that our measure can be relied upon to capture meaningful variations in answer quality. This experiment also defines reference points for interpreting cosine similarity values, though we must use them cautiously to avoid misleading comparisons.

Figure \ref{fig:validity_experiment} displays the results of this experiment, showing the average cosine similarity (on the left axis) and the observation count (on the right axis) for each reply category.\footnote{A figure illustrating the distribution of cosine similarity estimates for each reply category is included in the Online Supplementary Material.} The differences in average cosine similarity across all categories are statistically significant at the 95\% confidence level, implying that our measure of answer quality aligns with the previously defined taxonomy of replies while offering a complementary perspective.

The difference between the average cosine similarity for full replies (0.627) and non-replies (0.433) is small relative to the range of cosine similarity estimates across all pairs. In addition, the average cosine similarity for intermediate replies (0.545) is nearly equal to that for the entire inference set (0.539), suggesting that the typical answer falls into this intermediate category. Notably, the average cosine similarity for non-replies is significantly above zero and nears the upper bound of the null distribution, indicating that non-replies are generally more relevant than random answers.

\afterpage{\FloatBarrier}

\section{Hypotheses}

We now investigate the relationship between answer quality and relevant variables. This analysis seeks to understand the behavior of QP participants and identify factors associated with higher answer relevance. It also serves as a test of our measurement approach's face validity: if the expected correlations do not materialize, it may suggest that our methodology does not effectively capture answer quality.

We focus our analysis on the relationship between the quality of answers and power dynamics in the House of Commons. We propose three hypotheses:
\begin{enumerate}[label=(\roman*)]
    \item When the government has a minority of seats in the House of Commons, its answers tend to be more relevant.

    \item The relevance of answers varies with MPs' party affiliation, favoring government backbench members and those from parties with which the government is closer ideologically.

    \item The relevance of answers to questions from opposition parties varies with the size of their caucus.
\end{enumerate}

The first hypothesis posits that a minority government has more incentives to collaborate with opposition parties since its survival hinges on their support in confidence votes. In this context, inadequate answers to the opposition's questions risk alienating them and jeopardizing the government's stability.

The second hypothesis suggests that the LPC, for example, may provide higher-quality answers to questions from the NDP than those from the CPC. Additionally, government backbenchers asked 4,061 (7.4$\%$) questions in the inference set. These questions are often ``planted questions,'' prearranged, friendly queries meant to highlight the government's achievements or criticize the opposition rather than scrutinize its actions. We expect a high similarity between these questions and their answers.

The third hypothesis considers how the size of an opposition party's caucus influences the quality of government answers. On the one hand, a larger caucus could enhance the party's ability to secure high-quality answers by boosting its influence on confidence motions. On the other hand, a larger caucus might heighten competition with the government, prompting the opposition party to adopt a more aggressive questioning style and, in turn, cause the government to prioritize counterattacks over addressing the questions' substance, ultimately diminishing the quality of answers.

Note that our analysis does not establish a causal relationship between answer quality and variables of interest. It is also crucial to acknowledge the questions' endogeneity. Accordingly, the incentives of the government and opposition parties both influence answer quality. We must account for this factor in the interpretation of our findings.

\section{Results}

\begin{figure}[!p]
    \centering
    \includegraphics{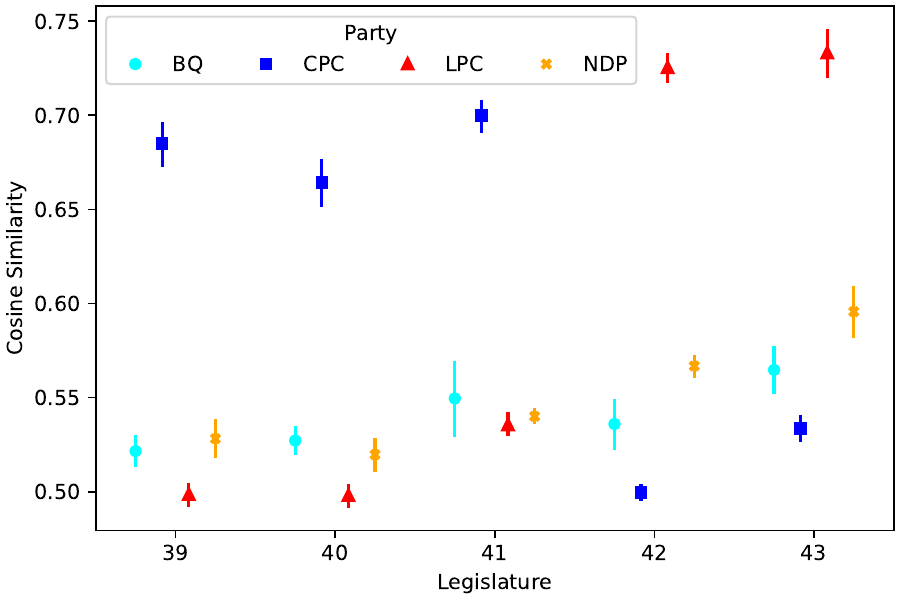}
    \caption{Average Cosine Similarity Between Questions and Answers by Party and Legislature}
    \label{fig:cosine_by_legislature}
\end{figure}

We now turn to the results of our analysis. Figure \ref{fig:cosine_by_legislature} illustrates the evolution of the average cosine similarity between questions and answers conditional on the party affiliation of the MP asking the question across the five legislatures within our period of interest. For context, the CPC held office between the 39\textsuperscript{th} and 41\textsuperscript{st} legislatures, whereas the LPC held office during the 42\textsuperscript{nd} and 43\textsuperscript{rd} legislatures. Also, minority governments held office during the 39\textsuperscript{th}, 40\textsuperscript{th}, and 43\textsuperscript{rd} legislatures.

\afterpage{\FloatBarrier}

We draw four conclusions from this figure. First, answers to questions from government backbenchers have, on average, much higher quality than those to questions from opposition members. The difference in the relevance of answers to questions from government backbenchers compared to opposition MPs is much greater than the difference in the relevance of answers to questions from members of various opposition parties. We expected this since members of the ruling party are unlikely to ask questions that would embarrass or hurt the government they belong to. As noted earlier, it is common for ministerial aides to arrange for government backbenchers to ask friendly questions. This finding gives credence to the belief that this is a common practice.

Second, there is mixed evidence regarding the hypothesized relationship between the quality of answers and whether the government holds a minority or majority of seats in the House of Commons. Comparing the 39\textsuperscript{th} and 40\textsuperscript{th} legislatures to the 41\textsuperscript{st}, it seems that the CPC offered, on average, more relevant answers to questions from the opposition when it held a majority of seats in the House of Commons, contrary to our hypothesis. On the other hand, comparing the 42\textsuperscript{nd} and 43\textsuperscript{rd} legislatures, we find that the LPC offered, on average, more relevant answers to questions from the opposition when it held a minority of seats in the House. However, the 43\textsuperscript{rd} legislature coincided with the COVID-19 pandemic, which likely affected the nature of the questions and the government's inclination to answer them transparently. Given this ambiguous evidence, we do not consider that the data confirms our hypothesis.

Third, third-party MPs receive, on average, more relevant answers than MPs from the official opposition. During the 39\textsuperscript{th}, 40\textsuperscript{th}, 42\textsuperscript{nd}, and 43\textsuperscript{rd} legislatures, the quality of answers to questions from members of third parties was significantly higher than the quality of answers to questions from members of the official opposition. The outcomes from the 41\textsuperscript{st} legislature, the first and only time the NDP formed the official opposition, further illustrate this. During that period, answers to members of the NDP were no different in quality from those given to members of the LPC. In contrast, there were statistically significant differences during the previous two legislatures. Figures in the Online Supplementary Material further support this finding, showing a statistically significant negative correlation between cosine similarity and the number and share of seats opposition parties hold. We attribute this difference to the fact that the media and public tend to see the official opposition as the ``government-in-waiting,'' positioning it in direct competition with the government. It can lead the official opposition to be more antagonistic and assertive in questioning the government and the latter to be more hostile and reticent in answering its questions.

Fourth, the difference between answers to questions from the official opposition and third parties is smaller than the variations in responses to questions from different opposition parties attributable to their perceived ideological proximity to the government. For example, during the 42\textsuperscript{nd} and 43\textsuperscript{rd} legislatures, questions from NDP members have received much more relevant answers, on average, than those from all opposition parties before and in the same period. This relationship was particularly pronounced during the 43\textsuperscript{rd} legislature when the LPC held a minority of seats in the House of Commons and heavily relied on the NDP to survive motions of confidence. Over that period, the representation of the NDP in the House of Commons was comparable to its representation during the 39\textsuperscript{th} and 40\textsuperscript{th} legislatures and the LPC's during the 41\textsuperscript{st} legislature. Therefore, we can impute this pattern to variations in the relative size of the NDP's caucus.

\section{Conclusion}

This paper proposes a novel approach for assessing answer quality, drawing inspiration from semantic search, a core task in information retrieval and NLP. Our methodology consists of measuring the quality of an answer by how easily and accurately it can be recognized from a random set of candidate answers given the question's text. This measure captures the relevance of answers to questions. It is primarily valuable for assessing the efficacy of QP as an accountability and oversight mechanism. A key advantage of our approach is that we can computationally implement it by fine-tuning a language model on the corpus of observed questions and answers without human-labeled data, making it highly efficient for analyzing large corpora.

We used our methodology to study the quality of answers during QP in the Canadian House of Commons. Our findings imply that we should not consider QP as a time when government ministers innocently deliver plain answers to every question. Nor should we dismiss it as a stage for opposition MPs and government ministers to engage in political messaging and deliver essentially independent speeches disguised as questions and answers. Indeed, while our analysis shows that some answers have a vague semantic connection to questions, hinting at some evasion or obfuscation, answers are generally moderately relevant to the questions asked, more so than they would if the government responded randomly. Therefore, in line with its stated purpose, QP effectively allows MPs to elicit insightful answers from the government.

Our measurement approach also allowed us to investigate correlates of answer quality. Our analysis revealed a correlation between the quality of answers and the party affiliation of the MP asking questions, with questions from government backbenchers, MPs from third parties (as opposed to the official opposition), and those from parties ideologically closer to the government receiving more relevant answers, on average. These findings stress the substantive value of our measurement approach and confirm its validity.

To complement this analysis, the Online Supplementary Material includes an analysis of the correlation between the quality of answers and the topics of the questions, as well as the results of two robustness checks. The first considers potential bias from sampling error in estimating latent representations. The second investigates the impact of including exchanges initiated by government backbenchers in the training set on our findings. The results confirm the robustness of our conclusions.

In conclusion, we can apply the measurement approach proposed in this article to many contexts within and beyond political science. Its applicability to institutions similar to QP, such as Prime Minister's Questions in the United Kingdom and \emph{Questions au Gouvernment} in France, is obvious. We could also use this approach to study parliamentary hearings, election debates, and press conferences, including in the United States. Beyond political science, this approach can contribute to analyzing news interviews with non-political public figures, press conferences by central bankers, or earning calls by executives of publicly traded companies. We encourage researchers to adopt this approach to study their institutions of interest and test their hypotheses. Our methodology may eventually require some adjustments and refinements. One area for further investigation is follow-up questions, especially in unmoderated or less structured settings. To accurately assess the quality of answers to those questions, it may be necessary to consider a broader context, including prior questions.

\clearpage

\section*{Funding}

This project was not financially supported by any public, commercial, or nonprofit organization.

\section*{Data Availability Statement}

Replication code and data for this article will be made available online prior to publication.

\section*{Supplementary Material}

Supplementary material will be made available online prior to publication.

\clearpage

\printbibliography

\end{document}